\documentclass[12pt, final]{l4dc2023}

\usepackage{amsmath,amsfonts,bm}

\newcommand{\figleft}{{\em (Left)}}

\newcommand{\figright}{{\em (Right)}}

\def\eqref#1{equation~\ref{#1}}

\def\1{\bm{1}}

\DeclareMathAlphabet{\mathsfit}{\encodingdefault}{\sfdefault}{m}{sl}
\SetMathAlphabet{\mathsfit}{bold}{\encodingdefault}{\sfdefault}{bx}{n}

\def\gD{{\mathcal{D}}}

\def\gL{{\mathcal{L}}}

\def\gS{{\mathcal{S}}}

\newcommand{\E}{\mathbb{E}}

\DeclareMathOperator*{\argmin}{arg\,min}

\usepackage{hyperref}
\usepackage{url}
\usepackage{amssymb}
\usepackage{bbm}
\usepackage{algorithm}
\usepackage{algpseudocode}
\usepackage{wrapfig}
\usepackage{graphicx}
\usepackage{caption}
\usepackage{enumitem}

\newtheorem{assumption}{Assumption}

\title[Contrastive Example-Based Control]{Contrastive Example-Based Control}
\usepackage{times}

\author{%
 \Name{Kyle Hatch}~$^{1}$ \Email{khatch@stanford.edu}\\
 \Name{Benjamin Eysenbach}~$^{2}$ \Email{beysenba@cs.cmu.edu}\\
 \Name{Rafael Rafailov}~$^1$ \Email{rafailov@stanford.edu}\\
 \Name{Tianhe Yu}~$^{1}$ \Email{tianheyu@cs.stanford.edu}\\
 \Name{Ruslan Salakhutdinov}~$^2$ 
 \Email{rsalakhu@cs.cmu.edu}\\
 \Name{Sergey Levine}~$^3$ 
 \Email{svlevine@eecs.berkeley.edu}\\
 \Name{Chelsea Finn}~$^1$ \Email{cbfinn@cs.stanford.edu}\\
 \addr $^1$Department of Computer Science, Stanford University \\
 \addr $^2$Machine Learning Department, Carnegie Mellon University \\
 \addr $^3$Department of Electrical Engineering and Computer Sciences, UC Berkeley
}
\begin{document}

\maketitle

\begin{abstract}%

While many real-world problems that might benefit from reinforcement learning, these problems rarely fit into the MDP mold: interacting with the environment is often expensive and specifying reward functions is challenging. Motivated by these challenges, prior work has developed data-driven approaches that learn entirely from samples from the transition dynamics and examples of high-return states. These methods typically learn a reward function from high-return states, use that reward function to label the transitions, and then apply an offline RL algorithm to these transitions. While these methods can achieve good results on many tasks, 
they can be complex, often requiring regularization and temporal difference updates.
In this paper, we propose a method for offline, example-based control that learns an implicit model of multi-step transitions, rather than a reward function.
We show that this implicit model can represent the Q-values for the example-based control problem.
Across a range of state-based and image-based offline control tasks, our method outperforms baselines that use learned reward functions; additional experiments demonstrate improved robustness and scaling with dataset size.%
\footnote{Videos of our method are available on the project website: 
\url{https://sites.google.com/view/laeo-rl}.
Code is released at: 
\url{https://github.com/khatch31/laeo}.}

\end{abstract}


\section{Introduction}
\label{sec:intro}

Reinforcement learning is typically framed as the problem of maximizing a given reward function. However, in many real-world situations, it is more natural for users to define what they want an agent to do with examples of successful outcomes~\citep{fu2018variational,zolna2020offline,xu2019positive,eysenbach2021replacing}. For example, a user that wants their robot to pack laundry into a washing machine might provide multiple examples of states where the laundry has been packed correctly. This problem setting is often seen as a variant of inverse reinforcement learning~\citep{fu2018variational}, where the aim is to learn only from examples of successful outcomes, rather than from expert demonstrations. To solve this problem, the agent must both figure out what constitutes task success (i.e., what the examples have in common) and how to achieve such successful outcomes.

In this paper, our aim is to address this problem setting in the case where the agent must learn from offline data without trial and error. Instead, the agent must infer the outcomes of potential actions from the provided data, while also relating these inferred outcomes to the success examples. We will refer to this problem of offline RL with success examples as \emph{offline example-based control}.

Most prior approaches involve two steps: \emph{first} learning a reward function, and \emph{second} combining it with an RL method to recover a policy~\citep{fu2018variational,zolna2020offline,xu2019positive}. While such approaches can achieve excellent results when provided sufficient data~\citep{kalashnikov2021mt, zolna2020offline}, learning the reward function is challenging when the number of success examples is small~\citep{li2021mural, zolna2020offline}. Moreover, these prior approaches are relatively complex (e.g., they use temporal difference learning) and have many hyperparameters.

Our aim is to provide a simple and scalable approach that avoids the challenges of reward learning. The main idea will be learning a certain type of dynamics model. Then, using that model to predict the probabilities of reaching each of the success examples, we will be able to estimate the Q-values for every state and action. Note that this approach does not use an offline RL algorithm as a subroutine. The key design decision is the model type; we will use an implicit model of the time-averaged future (precisely, the discounted state occupancy measure). This decision means that our model reasons across multiple time steps but will not output high-dimensional observations (only a scalar number).
A limitation of this approach is that it will correspond to a single step of policy improvement: the dynamics model corresponds to the dynamics of the behavioral policy, not of the reward-maximizing policy. While this means that our method is not guaranteed to yield the optimal policy, our experiments nevertheless show that our approach outperforms multi-step RL methods.

The main contribution of this paper is an offline RL method (LAEO) that learns a policy from examples of high-reward states. The key idea behind LAEO is an implicit dynamics model, which represents the probability of reaching states at some point in the future. We use this model to estimate the probability of reaching examples of high-return states. LAEO is simpler yet more effective than prior approaches based on reward classifiers. Our experiments demonstrate that LAEO can successfully solve offline RL problems from examples of high-return states on four state-based and two image-based manipulation tasks.
Our experiments show that LAEO is more robust to occlusions and also exhibits better scaling with dataset size than prior methods.
We show that LAEO can work in example-based control settings in which goal-conditioned RL methods fail. Additionally, we show that the dynamics model learned by LAEO can generalize to multiple different tasks, being used to solve tasks that are not explicitly represented in the training data.

\section{Related Work}
\label{sec:related_work}

\paragraph{Reward learning.} To overcome the challenge of hand-engineering reward functions for RL, prior methods either use supervised learning or adversarial training to learn a policy that matches the expert behavior given by the demonstration (imitation learning)~\citep{pomerleau1988alvinn,Dagger2011Ross,GAIL2016Ho,Feedback2021Spencer}
or learn a reward function from demonstrations and optimize the policy with the learned reward through trial and error (inverse RL)~\citep{ng2000irl, abbeel2004apprenticeship, ratliff2006, ziebart2008maximum, finn2016guided,AIRLFu2018}. 
However, providing full demonstrations complete with agent actions is often difficult, therefore, recent works have focused on the setting where only a set
of user-specified goal states or human videos are available~\citep{fu2018variational,singh2019end, kalashnikov2021mt, xie2018few, eysenbach2021replacing, chen2021learning}.
These reward learning approaches have shown successes in real-world robotic manipulation tasks from high-dimensional imageinputs~\citep{finn2016guided,singh2019end,zhu2020ingredients,chen2021learning}. Nevertheless, to combat covariate shift that could lead the policy to drift away from the expert distribution, these methods usually require significant online interaction. Unlike these works that study online settings, we consider learning visuomotor skills from offline datasets.

\paragraph{Offline RL.} Offline RL~\citep{ernst2005tree, riedmiller2005neural, LangeGR12, levine2020offline} studies the problem of learning a policy from a static dataset without online data collection in the environment, which has shown promising results in robotic manipulation~\citep{kalashnikov2018scalable, mandlekar2020iris, Rafailov2020LOMPO,singh2020cog,julian2020efficient,kalashnikov2021mt}. Prior offline RL methods focus on the challenge of distribution shift between the offline training data and deployment using a variety of techniques, such as policy constraints~ \citep{fujimoto2018off,liu2020provably,jaques2019way,wu2019behavior, zhou2020plas,kumar2019stabilizing,siegel2020keep, peng2019advantage, fujimoto2021minimalist,ghasemipour2021emaq}, conservative Q-functions~\citep{kumar2020conservative,kostrikov2021offline,yu2021combo, sinha2021s4rl}, and penalizing out-of-distribution states generated by learned dynamics models~\citep{kidambi2020morel, yu2020mopo,matsushima2020deployment,argenson2020model,swazinna2020overcoming,Rafailov2020LOMPO,lee2021representation,yu2021combo}. 

While these prior works successfully address the issue of distribution shift, they still require reward annotations for the offline data. Practical approaches have used manual reward sketching to train a reward model ~\citep{cabi2019framework, Konyushkova2020SSRewardLearning, Rafailov2020LOMPO} or heuristic reward functions ~\citep{yu2022UDS}. Others have considered offline learning from demonstrations, without access to a predefined reward function ~\citep{mandlekar2020iris,zolna2020offline, xu2022Discriminator, Jarboui2021OfflineIRL}, however they rely on high-quality demonstration data. 
In contrast, our method: \emph{(1)} addresses distributional shift induced by both the learned policy and the reward function in a principled way, \emph{(2)} only requires user-provided goal states and \emph{(3)} does not require expert-quality data, resulting in an effective and practical offline reward learning scheme.

\section{Learning to Achieve Examples Offline}
\label{sec:method}

Offline RL methods typically require regularization, and our method will employ regularization in two ways. First, we regularize the policy with an additional behavioral cloning term, which penalizes the policy for sampling out-of-distribution actions. Second, our method uses the Q-function for the behavioral policy, so it performs one (not many) step of policy improvement.  These regularizers mean that our approach is not guaranteed to yield the optimal policy. \looseness=-1

\subsection{Preliminaries}
\label{sec:prelims}

We assume that an agent interacts with an MDP with states $s \in \gS$, actions $a$, a state-only reward function $r(s) \ge 0$,
initial state distribution $p_0(s_0)$ and dynamics $p(s_{t+1} \mid s_t, a_t)$. We use $\tau = (s_0, a_0, s_1, a_1, \cdots)$ to denote an infinite-length trajectory. The likelihood of a trajectory under a policy $\pi(a \mid s)$ is $\pi(\tau) = p_0(s_0) \prod_{t=0}^\infty p(s_{t+1} \mid s_t, a_t) \pi(a_t \mid s_t)$. The objective is to learn a policy $\pi(a \mid s)$ that maximizes the expected, $\gamma$-discounted sum of rewards:
$\max_\pi \E_{\pi(\tau)}\left[\sum_{t=0}^\infty \gamma^t r(s_t)\right]. \label{eq:rl}$
We define the Q-function for policy $\pi$ as the expected discounted sum of returns, conditioned on an initial state and action:
\begin{equation}
    Q^\pi(s, a) \triangleq \E_{\pi(\tau)}\left[\sum_{t=0}^\infty \gamma^t r(s_t) \bigg \vert \substack{s_0 = s\\a_0 = a}\right]. \label{eq:q-vals}
\end{equation}
We will focus on the offline (i.e., batch RL) setting. Instead of learning by interacting with the environment (i.e., via trial and error), the RL agent will receive as input a dataset of trajectories $\gD_\tau = \{\tau \sim \beta(\tau) \}$ collected by a behavioral policy $\beta(a \mid s)$. We will use $Q^\beta(s, a)$ to denote the Q-function of the behavioral policy.

\paragraph{Specifying the reward function.}
In many real-world applications, specifying and measuring a scalar reward function is challenging, but providing examples of good states (i.e., those which would receive high rewards) is straightforward. Thus, we follow prior work~\citep{fu2018variational, zolna2020offline, eysenbach2021replacing, xu2019positive, zolna2020taskrelevant} in assuming that the agent does not observe scalar rewards (i.e., $\gD_\tau$ does not contain reward information). Instead, the agent receives as input a dataset $\gD_* = \{s^*\}$ of high-reward states $s^* \in \gS$. These high-reward states are examples of good outcomes, which the agent would like to achieve. The high-reward states are not labeled with their specific reward value.

To make the control problem well defined, we must relate these success examples to the reward function. We do this by assuming that the frequency of each success example is proportional to its reward: good states are more likely to appear (and be duplicated) as success examples.
\begin{assumption} \label{as:1}
Let $p_\tau(s)$ be the empirical probability density of state $s$ in the trajectory dataset, and let $p_*(s)$ as the empirical probability density of state $s$ under the high-reward state dataset. We assume that there exists a positive constant $c$ such that $r(s) = c \frac{p_*(s)}{p_\tau(s)}$ for all states $s$.
\end{assumption}
This is the same assumption as~\citet{eysenbach2021replacing}.
This assumption is important because it shows how example-based control is universal: for any reward function, we can specify the corresponding example-based problem by constructing a dataset of success examples that are sampled according to their rewards. We assumed that rewards are non-negative so that these sampling probabilities are positive. \looseness=-1

This assumption can also be read in reverse. When a user constructs a dataset of success examples in an arbitrary fashion, they are implicitly defining a reward function. In the tabular setting, the (implicit) reward function for state $s$ is the count of the times $s$ occurs in the dataset of success examples. 
Compared with goal-conditioned RL~\citep{kaelbling1993learning}, defining tasks via success examples is more general. By identifying what all the success examples have in common (e.g., laundry is folded), the RL agent can learn what is necessary to solve the task and what is irrelevant (e.g., the color of the clothes in the laundry). 
We now can define our problem statement as follows:
\begin{definition}
In the \textbf{offline example-based control} problem, a learning algorithm receives as input a dataset of trajectories $\gD_\tau = \{\tau\}$ and a dataset of successful outcomes $\gD_* = \{s\}$ satisfying Assumption~\ref{as:1}. The aim is to output a policy that maximizes the RL objective (Eq.~\ref{eq:rl}).
\end{definition}
This problem setting is appealing because it mirrors many practical RL applications: a user has access to historical data from past experience, but collecting new experience is prohibitively expensive. Moreover, this problem setting can mitigate the challenges of reward function design.
Rather than having to implement a reward function and add instruments to measure the corresponding components, the users need only provide a handful of observations that solved the task. This problem setting is similar to imitation learning, in the sense that the only inputs are data. However, unlike imitation learning, in this problem setting the high-reward states are not labeled with actions, and these high-reward states may not necessarily contain entire trajectories.

\begin{wrapfigure}[13]{R}{0.325\textwidth}
\centering
\includegraphics[width=0.9\linewidth]{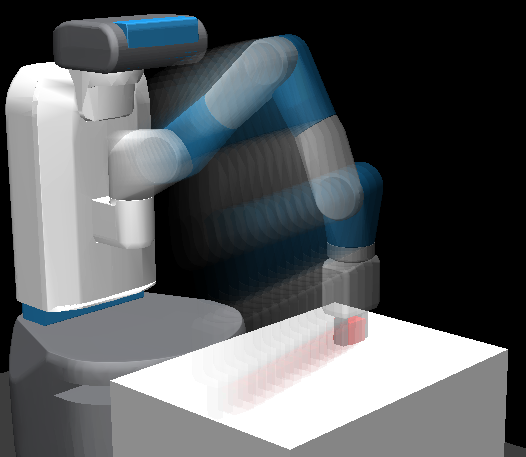} 
\caption{\footnotesize Our method will use contrastive learning to predict which states might occur at some point in the future.}
\label{fig:gamma-model}
\end{wrapfigure}

Our method will estimate the discounted state occupancy measure,
\begin{align}
    p^\beta(s_{t+} = s \mid s_0, a_0) &\triangleq (1 - \gamma) \sum_{t=0}^\infty \gamma^t p_t^\pi(s_t = s \mid s_0, a_0),
\end{align}
where $p_t^\beta(s_t \mid s, a)$ is the probability of policy $\beta(a \mid s)$ visiting state $s_t$ after exactly $t$ time steps.
Unlike the transition function $p(s_{t+1} \mid s_t, a_t)$, the discounted state occupancy measure indicates the probability of visiting a state at any point in the future, not just at the immediate next time step.
In tabular settings, this distribution corresponds to the successor representations~\citep{dayan1993improving}. To handle continuous settings, we will use the contrastive approach from recent work~\citep{mazoure2020deep, eysenbach2022contrastive}. We will learn a function $f(s, a, s_f) \in \mathbbm{R}$ takes as input an initial state-action pair as well as a candidate future state, and outputs a score estimating the likelihood that $s_f$ is a real future state. The loss function is a standard contrastive learning loss(e.g.,~\citet{ma2018noise}), where positive examples are triplets of a state, action, and future state:
\begin{equation*}
    \max_f \gL(f; \gD_\tau) \triangleq \E_{p(s, a), s_f \sim p^\beta(s_{t+} \mid s, a)}\left[\log \sigma(f(s, a, s_f)) \right] + \E_{p(s, a), s_f \sim p(s)}\left[\log (1 - \sigma(f(s, a, s_f))) \right],
\end{equation*}
where $\sigma(\cdot)$ is the sigmoid function. At optimality, the implicit dynamics model encodes the discounted state occupancy measure:
\begin{equation}
    f^*(s, a, s_f) = \log p^\beta(s_{t+} = s_f \mid s, a) - \log p_\tau(s_f).
\end{equation}

We visualize this implicit dynamics model in Fig.~\ref{fig:gamma-model}. Note that this dynamics model is policy dependent. Because it is trained with data collected from one policy ($\beta(a \mid s)$), it will correspond to the probability that \emph{that} policy visits states in the future. Because of this, our method will result in estimating the value function for the behavioral policy (akin to 1-step RL~\citep{brandfonbrener2021offline}), and will not perform multiple steps of policy improvement. Intuitively, the training of this implicit model resembles hindsight relabeling~\citep{kaelbling1993learning, andrychowicz2017hindsight}.  However, it is generally unclear how to use hindsight relabeling for single-task problems. Despite being a single-task method, our method will be able to make use of hindsight relabeling to train the dynamics model.

\subsection{Deriving Our Method}

The key idea behind out method is that this implicit dynamics model can be used to represent the Q-values for the example-based problem, up to a constant. The proof is in Appendix~\ref{appendix:proofs}.

\begin{lemma}
Assume that the implicit dynamics model is learned without errors. Then the Q-function for the data collection policy $\beta(a \mid s)$ can be expressed in terms of this implicit dynamics model:
\begin{align}
    Q^\beta(s, a)
    &= \frac{c}{1-\gamma} \E_{p_*(s^*)}\left[ e^{f(s, a, s^*)} \right].
\end{align}
\end{lemma}

So, after learning the implicit dynamics model, we can estimate the Q-values by averaging this model's predictions across the success examples. We will update the policy using Q-values estimated in this manner, plus a regularization term:
{\begin{equation}
    \min_\pi \gL(\pi; f, \gD_*) \triangleq -(1 - \lambda) \E_{\pi(a \mid s)p(s), s^* \sim \gD_*}\left[e^{f(s, a, s^*)} \right] - \lambda \E_{s, a \sim \gD_\tau}\left[\log \pi(a \mid s) \right]. \label{eq:critic-loss}
\end{equation}
In our experiments, we use a weak regularization coefficient of $\lambda = 0.5$.

\begin{wrapfigure}[17]{r}{0.5\textwidth}
    \centering
    \vspace{-1.2em}
    \includegraphics[width=\linewidth]{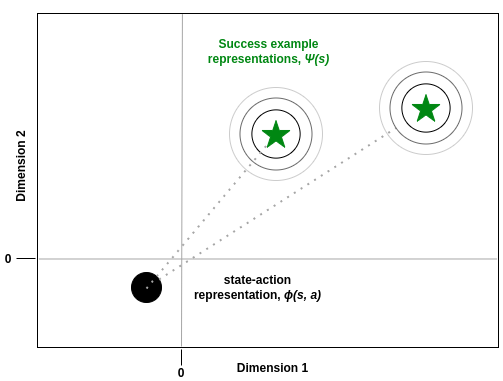}%
    \vspace{-0.5em}%
    \caption{\footnotesize 
    If the state-action representation $\phi(s, a)$ is close to the representation of a high-return state $\psi(s)$, then the policy is likely to visit that state. Our method estimates Q-values by combining the distances to all the high-return states (Eq.~\ref{eq:q-vals}).
    }
    \label{fig:geometry}
\end{wrapfigure}

It is worth comparing this approach to prior methods based on learned reward functions~\citep{xu2019positive, fu2018variational, zolna2020offline}. Those methods learn a reward function from the success examples, and use that learned reward function to synthetically label the dataset of trajectories. Both approaches can be interpreted as learning a function on one of the datasets and then applying that function to the other dataset.
Because it is easier to fit a function when given large quantities of data, we predict that our approach will outperform the learned reward function approach when the number of success examples is small, relative to the number of unlabeled trajectories.
Other prior methods~\citep{eysenbach2021replacing,SQIL2020Reddy} avoid learning reward functions by proposing TD update rules that are applied to both the unlabeled transitions and the high-return states. However, because these methods have yet to be adapted to the offline RL setting, we will focus our comparisons on the reward-learning methods.

\subsection{A Geometric Perspective}

Before presenting the complete RL algorithm, we provide a geometric perspective on the representations learned by our method. Our implicit models learns a representation of state-action pairs $\phi(s, a)$ as well as a representation of future states $\psi(s)$. One way that our method can optimize these representations is by treating $\phi(s, a)$ as a prediction for the future representations.\footnote{Our method can also learn the opposite, where $\psi(s)$ is a prediction for the previous representations.} Each of the high-return states can be mapped to the same representation space. To determine whether a state-action pair has a large or small Q-value, we can simply see whether the predicted representation $\phi(s, a)$ is close to the representations of any of the success examples. Our method learns these representations so that the Q-values are directly related to the Euclidean distances\footnote{When representations are normalized, the dot product is equivalent to the Euclidean norm. We find that unnormalized features work better in our experiments.} from each success example. Thus, our method can be interpreted as learning a representation space such that estimating Q-values corresponds to simple geometric operations (kernel smoothing with an RBF kernel~\citep[Chpt.~6]{hastie2009elements}) on the learned representations. While the example-based control problem is more general than goal-conditioned RL (see Sec.~\ref{sec:prelims}), we can recover goal-conditioned RL as a special case by using a single success example.

\subsection{A Complete Algorithm}

We now build a complete offline RL algorithm based on these Q-functions.
We will call our method \textsc{Learning to Achieve Examples Offline} (LAEO).
Our algorithm will resemble one-step RL methods, but differ in how the Q-function is trained. After learning the implicit dynamics model (and, hence, Q-function) we will optimize the policy. The objective for the policy is maximizing (log) Q-values plus a %
regularization term, which penalizes sampling unseen actions:\footnote{For all experiments except Fig.~\ref{fig:examplebased}, we apply Jensen's inequality to the first term, using $\E_{\pi(a \mid s), s^* \sim p_*(s)}[f(s, a, s^*)]$.}
\begin{align}
    \max_\pi \; & (1 - \lambda) \log \E_{\pi(a \mid s)p_\tau(s)}\left[Q(s, a) \right] + \lambda \E_{(s, a) \sim p_\tau(s, a)}\left[\log \pi(a \mid s) \right] \nonumber \\
    &=  (1 - \lambda) \log \E_{\pi(a \mid s), s^* \sim p_*(s)}\left[e^{f(s, a, s^*)} \right] + \lambda \E_{(s, a) \sim p_\tau(s, a)}\left[\log \pi(a \mid s) \right]. \label{eq:policy-loss}
\end{align}

\begin{wrapfigure}[8]{R}{0.5\textwidth}
\vspace{-1.2em}
\begin{minipage}{0.5\textwidth}
\begin{algorithm}[H] \footnotesize
    \caption{{\small Learning to Achieve Examples Offline}}\label{alg:method}
    \begin{algorithmic}[1]
    \State \textbf{Inputs}: dataset of trajectories $\gD = \{\tau\}$, \phantom{.............} dataset of high-return states $\gD_* = \{s\}$.
    \State Learn the model via contrastive learning: \phantom{...............} $f \gets \argmin_f \gL(f; \gD_\tau)$ \Comment{Eq.~\ref{eq:critic-loss}}
    \State Learn the policy: $\pi \gets \argmin_\pi \gL(\pi; f, \gD_*)$ \Comment{Eq.~\ref{eq:policy-loss}}
    \State \textbf{return} policy $\pi(a \mid s)$
    \end{algorithmic}
\end{algorithm}
\end{minipage}
\end{wrapfigure}

As noted above, this is a one-step RL method: it updates the policy to maximize the Q-values of the behavioral policy.
Performing just a single step of policy improvement can be viewed as a form of regularization in RL, in the same spirit as early stopping is a form of regularization in supervised learning. Prior work has found that one-step RL methods can perform well in the offline RL setting. Because our method performs only a single step of policy improvement, we are not guaranteed that it will converge to the reward-maximizing policy. We summarize the complete algorithm in Alg.~\ref{alg:method}.

\section{Experiments}
\label{sec:experiments}

\begin{wrapfigure}[11]{R}{0.7\textwidth}
    \vspace{-0.7em}
    \centering
    \includegraphics[width=0.24\linewidth]{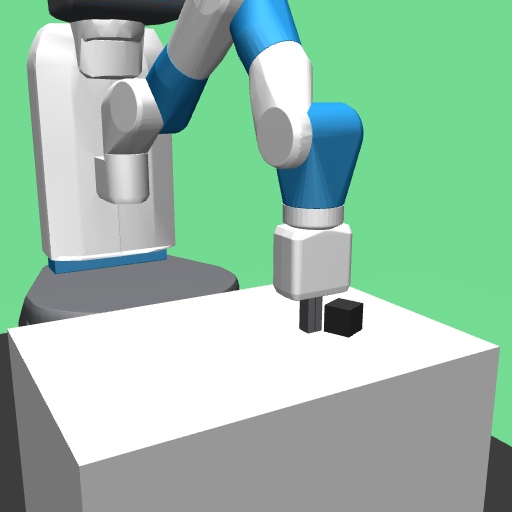}%
    \includegraphics[width=0.24\linewidth]{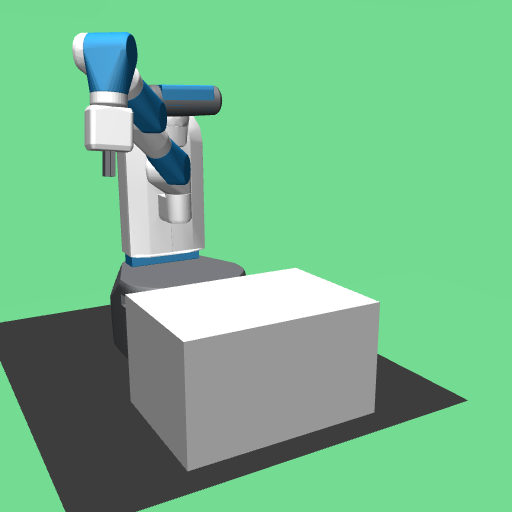}
    \includegraphics[width=0.24\linewidth]{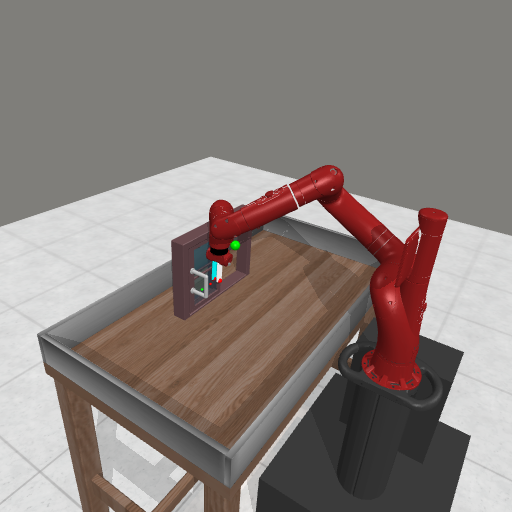}%
    \includegraphics[width=0.24\linewidth]{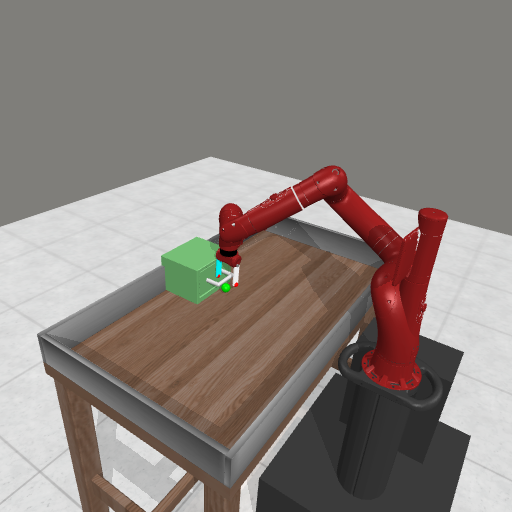}
    \caption{\footnotesize \textbf{Benchmark tasks}: We evaluate the performance of LAEO on six simulated manipulation tasks, two of which use pixel observations (\texttt{FetchReach-image} and  \texttt{FetchPush-image}) and four of which use low-dimensional states (\texttt{FetchReach}, \texttt{FetchPush}, \texttt{SawyerWindowOpen}, and \texttt{SawyerDrawerClose} ). }
    \label{fig:main_tasks}
\end{wrapfigure}

Our experiments test whether LAEO can effectively solve offline RL tasks that are specified by examples of high-return states, rather than via scalar reward functions. We study when our approach outperforms prior approaches based on learned reward functions. We look not only at the performance relative to baselines on state-based and image-based tasks, but also how that performance depends on the size and composition of the input datasets. 
Additional experiments study how LAEO performs when provided with varying numbers of success observations and whether our method can solve partially observed tasks. 
We include full hyperparameters and implementation details in Appendix \ref{appendix:details}. 
Code is available at \url{https://github.com/khatch31/laeo}.
Videos of our method are available at \url{https://sites.google.com/view/laeo-rl}.

\paragraph{Baselines.}
Our main point of comparison will be prior methods that use learned reward functions:
ORIL~\citep{zolna2020offline} and 
PURL~\citep{xu2019positive}. 
The main difference between these methods is the loss function used to train reward function: ORIL uses 
binary cross entropy loss 
while PURL uses a positive-unlabeled loss~\citep{xu2019positive}. 
Note that the ORIL paper also reports results using a positive-unlabeled loss, but for the sake of clarity we simply refer to it as PURL.
After learning the reward function, each of these methods applies an off-the-shelf RL algorithm. We will implement all baselines using the TD3+BC~\citep{fujimoto2021minimalist} offline RL algorithm. These offline RL methods achieve good performance on tasks specified via reward functions~\citep{kostrikov2021offline, brandfonbrener2021offline, fujimoto2021minimalist}.
We also include Behavioral Cloning (BC) results.

\begin{figure}[t]
    \vspace{-0.7em}
    \centering
    \includegraphics[width=\linewidth]{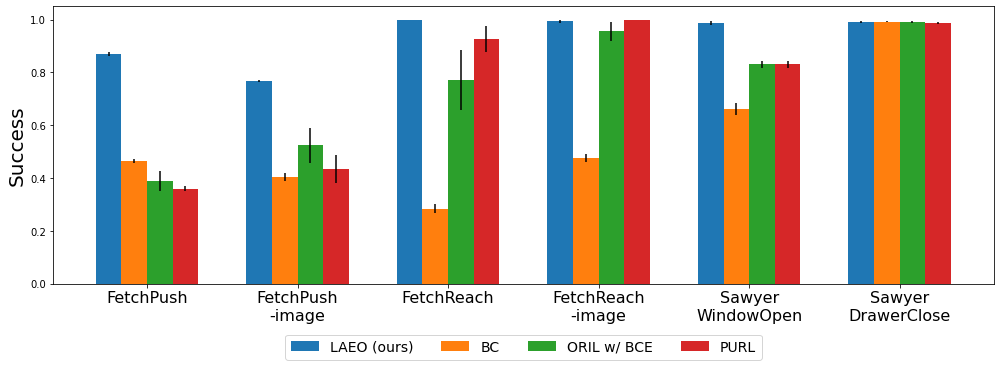}
    \caption{\footnotesize \textbf{Benchmark comparison}: 
    LAEO matches  or outperforms prior example-based offline RL methods on state and image-based tasks, including those that learn a separate reward function (ORIL, PURL).The gap in performance is most significant on the \texttt{FetchPush} and \texttt{FetchPush-image} tasks, which involve more complicated dynamics than the other tasks, suggesting that LAEO may outperform model free reward-learning approaches on tasks with complicated dynamics. 
    LAEO also outperforms BC on all of the tasks, highlighting LAEO's ability to learn a policy that outperforms the behavior policy on non-demonstration datasets.
    }
    \label{fig:main_results}
    \vspace{-0.5em}
\end{figure}

\paragraph{Benchmark comparison.}

We start by comparing the performance of LAEO to these baselines on six manipulation tasks. \texttt{FetchReach} and \texttt{FetchPush} are two manipulation tasks from~\citet{plappert2018multi} that use state-based observations. \texttt{FetchReach-image} and \texttt{FetchPush-image} are the same tasks but with image-based observations. \texttt{SawyerWindowOpen} and 
\texttt{Sawyer-} \texttt{DrawerClose}
are two manipulation tasks from~\citet{yu2020metaworld}.
For each of these tasks, we collect a dataset of medium quality by training an online agent from~\citet{eysenbach2022contrastive} and rolling out multiple checkpoints during the course of training. 
The resulting datasets have success rates between  $45\% - 50\%$.
We report results after $500,000$ training gradient steps (or $250,000$ steps, if the task success rates have converged by that point). \looseness=-1

We report results in Fig.~\ref{fig:main_results}.
We observe that LAEO, PURL, and ORIL perform similarly on  \texttt{FetchReach} and \texttt{FetchReach-image}. This is likely because these are relatively easy tasks, and each of these methods is able to achieve a high success rate. Note that all of these methods significantly outperform BC, indicating that they are able to learn better policies than the mode behavior policies represented in the datasets. 
On \texttt{SawyerDrawerClose}, all methods, including BC, achieve near perfect success rates, likely due to the simplicity of this task.
On \texttt{FetchPush}, \texttt{FetchPush-image}, and \texttt{SawyerWindowOpen}, LAEO outperforms all of the baselines by a significant margin. 
Recall that the main difference between LAEO and 
PURL/ORIL
is by learning a dynamics model, rather than the reward function.  These experiments suggest that 
for tasks with more complex dynamics, learning a dynamics model can achieve better performance than is achieved by model-free reward classifier methods.

\begin{wrapfigure}[10]{R}{0.5\textwidth}
    \vspace{-0.9em}
    \centering
    \includegraphics[height=3cm]{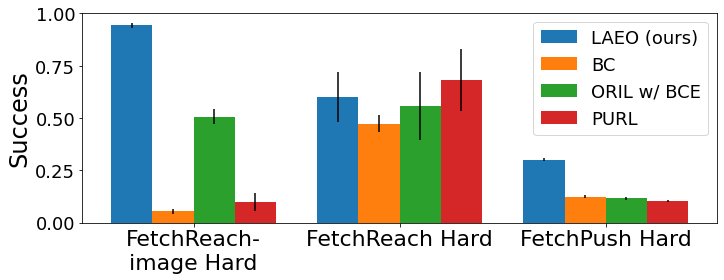}
    \vspace{-1.8em}
    \caption{\footnotesize \textbf{Data quality.} LAEO continues to match or outperform reward classifier based methods on datasets that contain a low percentage of successful trajectories.}
    \label{fig:hard-results}
\end{wrapfigure}

\paragraph{Varying the input data.}

Our next experiment studies how the dataset composition affects LAEO and the baselines. 
On each of three tasks, we generate a low-quality dataset by rolling out multiple checkpoints from a partially trained agent from~\citet{eysenbach2022contrastive}.
In comparison to the medium-quality datasets collected earlier, which have success rates between $45\% - 50\%$, these low quality datasets have success rates between $8\%-12\%$.
We will denote these low quality datasets with the ``Hard'' suffix.
Fig.~\ref{fig:hard-results} shows that LAEO continues to outperform baselines on these lower-quality datasets.

\begin{figure}[b]
    \centering
    \includegraphics[height=3.2cm]{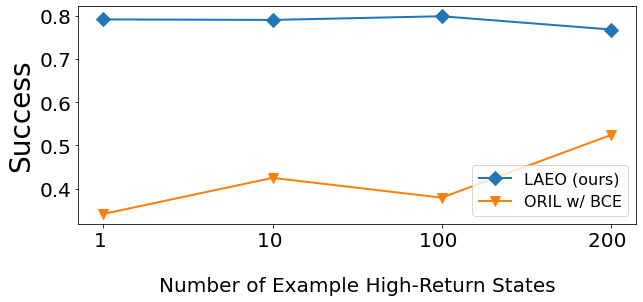}
    \hfill
    \includegraphics[height=3.3cm]{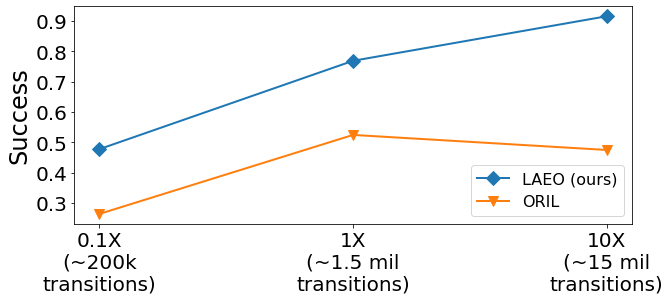}
    \caption{\footnotesize \textbf{Effect of dataset size}: 
    \figleft \, The most competitive baseline (ORIL) achieves better performance when given more examples of high-return states, likely because it makes it easier to learn ORIL's reward classifier. LAEO, which does not require learning a reward classifier, consistently achieves high success rates.
    \figright \, LAEO continues to improve when trained with more reward-free trajectories, while ORIL's performance plateaus. }
    \label{fig:data}
\end{figure}

Our next experiments study how varying the number of high-return example states and the number of reward-free trajectories affects performance. As noted in the Sec.~\ref{sec:intro}, we conjecture that our method will be especially beneficial relative to reward-learning approaches in settings with very few high-return example states.
In Fig.~\ref{fig:data} \emph{(left)}, 
we vary the number of high-return example states on 
\texttt{FetchPush} \texttt{-image}, 
holding the number of unlabeled trajectories constant. We observe that LAEO maintains achieves the same performance with 1 success example as with 200 success examples. In contrast, ORIL's performance decreases as the number of high-return example states decreases.
In Fig.~\ref{fig:data} \emph{(right)}, we vary the number of unlabeled trajectories, holding the number of high-return example states constant at $200$. 
We test the performance of LAEO vs. ORIL on three different dataset sizes on \texttt{FetchPush-image}, roughly corresponding to three different orders of magnitude: the $0.1\times$ dataset contains $3,966$ trajectories, %
the $1\times$ dataset contains $31,271$ trajectories, %
and the $10\times$ dataset contains $300,578$ trajectories. %
We observe that LAEO continues to see performance gains as number of unlabeled trajectories increases, whereas ORIL's performance plateaus.
Taken together these results suggest that, in comparison to reward classifier based methods, LAEO needs less human supervision and is more effective at leveraging large quantities of unlabeled data.

\paragraph{Partial Observability.}

\begin{wrapfigure}[13]{R}{0.5\textwidth}
    \centering
    \includegraphics[height=0.42\linewidth]{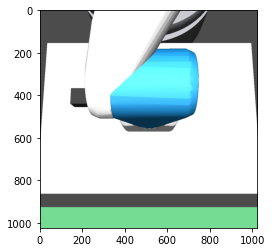}%
    ~
    \includegraphics[height=0.42\linewidth]{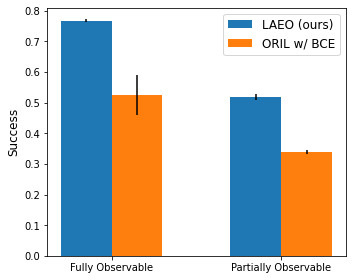}
    \vspace{-1.6em}
    \caption{\footnotesize \textbf{Partial observability.} LAEO continues to solve the \texttt{FetchPush-image} manipulation task in a setting where the new camera placement causes partial observability. This camera angle causes the block to be hidden from view by the gripper when the gripper reaches down to push the block. }
    \label{fig:partial}
\end{wrapfigure}

We also test the performance of LAEO on a partially-observed task. We modify the camera position in the \texttt{FetchPush-image} so that the block is occluded whenever the end effector is moved to touch the block. While such partial observability can stymie temporal difference methods~\citep{whitehead1991learning}, we predict that LAEO might continue to solve this task because it does not rely on temporal difference learning. 
The results, shown in Fig.~\ref{fig:partial}, confirm this prediction. 
On this partially observable task, we compare the performance of LAEO with that of ORIL, the best performing baseline on the fully observable tasks. On the partially observable task, LAEO achieves a 
success
rate of $51.9\%$, versus $33.9\%$ for ORIL.

\begin{wrapfigure}{R}{0.5\textwidth}
    \centering
    \vspace{-0.8em}
    \includegraphics[width=\linewidth]{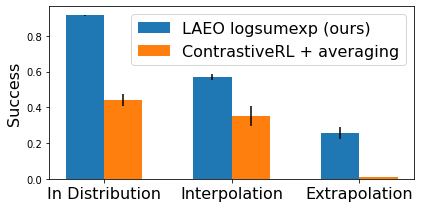}%
    \vspace{-0.5em}
    \caption{\footnotesize 
    \textbf{Comparison with goal-conditioned RL.} 
    LAEO solves manipulation tasks at multiple different locations 
    without being provided with a goal-state at test time. 
    }
    \label{fig:examplebased}
\end{wrapfigure}

\paragraph{Comparison to Goal-Conditioned RL.}
\label{par:comparison_to_goal_conditioned}

One of the key advantages of example-based control, relative to goal-conditioned RL, is that the policy can identify common patterns in the success examples to solve tasks in scenarios where it has never before seen a success example.
In settings such as robotics, this can be an issue since acquiring a goal state to provide to the agent requires already solving the desired task in the first place.
We test this capability in a variant of the \texttt{SawyerDrawerClose} environment. For training, the drawer's X position is chosen as one of five fixed locations. Then, we evaluate the policy learned by LAEO on three types of environments: \emph{In Distribution}: the drawer's X position is one of the five locations from training; \emph{Interpolation}: The drawer's X position is between some of the locations seen during training; \emph{Extrapolation}: The drawer's X position is outside the range of X positions seen during training. We compare to a goal-conditioned policy learned via contrastive RL, where actions are extracted by averaging over the (training) success examples: $\pi(a \mid s) = \E_{s^* \sim p_*(s)}[\pi(a \mid s, g=s^*)]$.

The results, shown in Fig.~\ref{fig:examplebased}, show that LAEO consistently outperforms this goal-conditioned baseline. As expected, the performance is highest for the In Distribution environments and lowest for the Extrapolation environments. 
Taken together, these experiments show that LAEO can learn to reach multiple different goal locations without access to goal states during test time. \looseness=-1

\paragraph{Multitask Critic.}

\begin{wrapfigure}[18]{R}{0.5\textwidth}
    \centering
    \includegraphics[width=\linewidth]{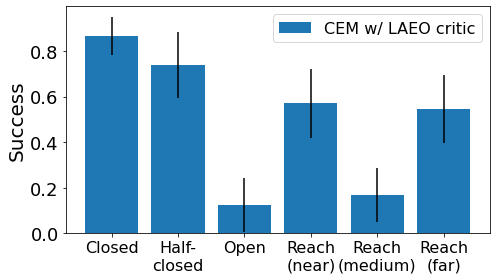}
    \caption{\footnotesize \textbf{Multitask Critic}: 
    Cross entropy method (CEM) optimization over the LAEO dynamics model trained only on the data from the drawer close task is able to solve six different tasks. Randomly sampling actions from the action space results in a $0\%$ success rate across all of the six tasks (not shown for clarity).}
    \label{fig:multitask_results}
\end{wrapfigure}

We explore whether a LAEO dynamics network trained on data from one task can be used to solve other downstream tasks.
We create a simple multitask environment by defining several different tasks that can be solved in the \texttt{SawyerDrawerClose} environment: \texttt{Close}, \texttt{Half-closed}, \texttt{Open}, \texttt{Reach-near}, \texttt{Reach-medium}, and \texttt{Reach-far}.
We then use a trained critic network from the previous set of experiments (Comparison to Goal-Conditioned RL), condition it on a success example from a downstream task, and select actions by using cross entropy method (CEM) optimization. By using CEM optimization, we do not need to train a separate policy network for each of the tasks. See Appendix \ref{appendix:multitask} for implementation details and for details of the multitask drawer environment.

CEM over the LAEO critic achieves non-zero success rates on all six tasks, despite only being trained on data from the \texttt{Close} task (see Figure \ref{fig:multitask_results}). In contrast, randomly sampling actions from the action space achieves a $0\%$ success rate on all of the tasks. 
Results are averaged across eight random seeds.
This suggests that a single LAEO critic can be leveraged to solve multiple downstream tasks, as long as the dynamics required to solve those tasks are represented in the training data.
Note that since we condition the critic network on a single goal example, these experiments can be interpreted from a goal-conditioned perspective as well as an example-based control perspective. In future work, we aim to explore the multitask capabilities of the LAEO dynamics model in an example-based control setting at a larger scale. This will involve training on larger, more diverse datasets as well as conditioning the critic network on multiple success examples for a single task (as done in the Comparison to Goal-Conditioned RL experiments).

\section{Conclusion}

In this paper, we present an RL algorithm aimed at settings where data collection and reward specification are difficult. Our method learns from a combination of high-return states and reward-free trajectories, integrating these two types of information to learn reward-maximizing policies. Whereas prior methods perform this integration by learning a reward function and then applying an off-the-shelf RL algorithm, ours learns an implicit dynamics model. Not only is our method simpler (no additional RL algorithm required!), but also it achieves higher success rates than prior methods.

While our experiments only start to study the ability of contrastive-based methods to scale to high-dimensional observations, we conjecture that methods like LAEO may be particularly amenable to such problems because the method for learning the representations (contrastive learning) resembles prior representation learning methods~\citep{mazoure2020deep, nair2022r3m}. Scaling this method to very large offline datasets is an important direction for future work.  \looseness=-1

\section{Acknowledgments}

BE is supported by the Fannie and John Hertz Foundation and the NSF GRFP (DGE2140739).


\newpage
\appendix
\section{Proofs}
\label{appendix:proofs}

The proof follows by substituting Assumption~\ref{as:1} into the definition of Q-values (Eq.~\ref{eq:q-vals}):
\begin{proof}
{\footnotesize \begin{align*}
    Q^\beta(s, a)
    = \E_{\beta(\tau)}\left[\sum_{t=0}^\infty \gamma^t r(s_t) \bigg \vert \substack{s_0 = s\\a_0 = a}\right] &= \frac{1}{1-\gamma} \int p^\beta(s_{t+} = s^* \mid s, a) r(s^*) ds^* \\
    &= \frac{1}{1-\gamma} \int p^\beta(s_{t+} = s^* \mid s, a)  c \frac{p_*(s^*)}{p_\tau(s^*)} ds^* \\
    &= \frac{c}{1-\gamma} \int p_*(s^*) e^{f(s, a, s^*)} ds^* = \frac{c}{1-\gamma} \E_{s^* \sim p_*(s)}\left[ e^{f(s, a, s^*)} \right].
\end{align*}}
\end{proof}

\section{Experimental Details}
\label{appendix:details}

We implemented our method and all baselines using the ACME framework~\citep{hoffman2020acme}. 
\begin{itemize}
    \item Batch size: 1024 for state based experiments, 256 for image based experiments
    \item Training iterations: $250,000$ if task success rates had converged by that point, otherwise $500,000$
    \item Representation dimension: 256
    \item Reward learning loss (for baselines): binary cross entropy (for ORIL) and positive unlabeled (for PURL)
    \item Critic architecture: Two-layer MLP with hidden sizes of 1024. ReLU activations used between layers. 
    \item Reward function architecture (for baselines): Two-layer MLP with hidden sizes of 1024. ReLU activations used between layers. 
    \item Actor learning rate: $3 \times 10^{-4}$
    \item Critic learning rate: $3 \times 10^{-4}$
    \item Reward learning rate (for baselines): $1 \times 10^{-4}$
    \item $\lambda$ for behavioral cloning weight in policy loss term: $0.5$
    \item $\eta$ for PU loss: $0.5$
    \item Size of offline datasets: Each dataset on \texttt{Fetch} tasks the consists of approximately $4,000$ trajectories of length $50$, except for the \texttt{FetchPush-image} dataset, which consists of approximately $40,000$ trajectories. Each dataset on the \texttt{Sawyer} tasks consists of approximately $4,000$ trajectories of length $200$.
\end{itemize}

\section{Multitask Critic Experiments}
\label{appendix:multitask}

\begin{figure}[t]
    \vspace{-0.7em}
    \centering
    \includegraphics[width=0.3\linewidth]{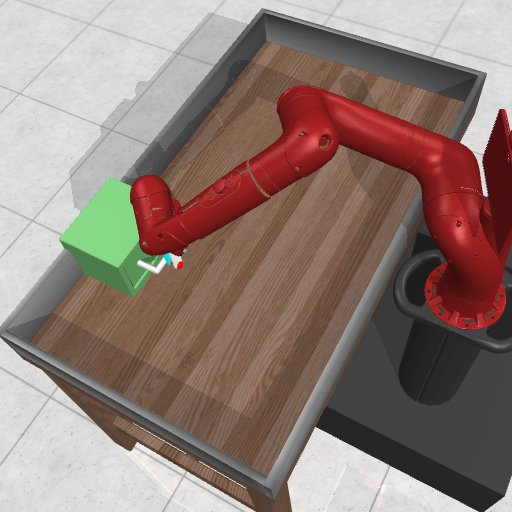}%
    ~
    \includegraphics[width=0.3\linewidth]{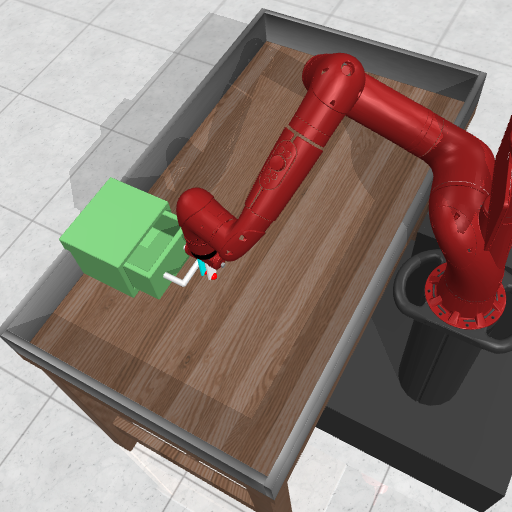}
    ~
    \includegraphics[width=0.3\linewidth]{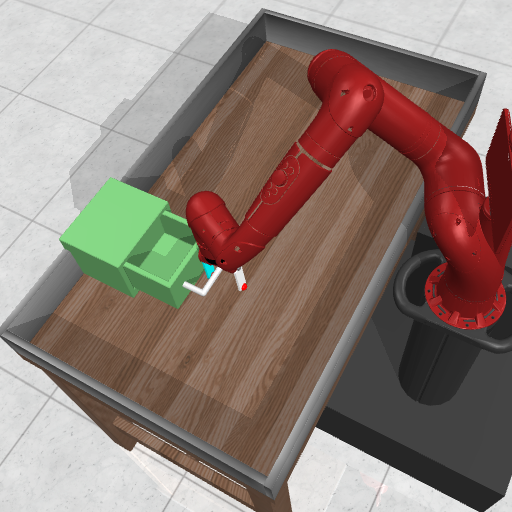}%
    
    \includegraphics[width=0.3\linewidth]{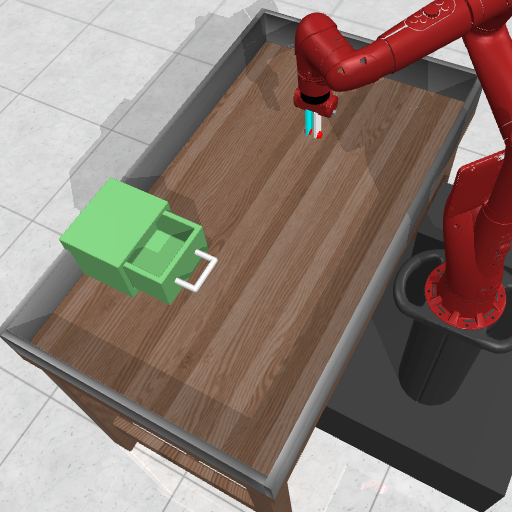}
    ~
    \includegraphics[width=0.3\linewidth]{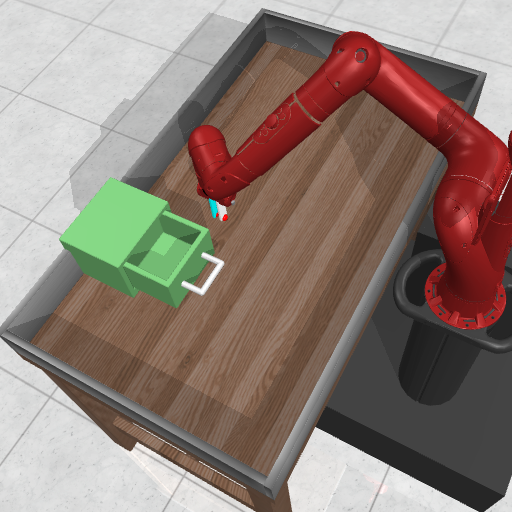}
    ~
    \includegraphics[width=0.3\linewidth]{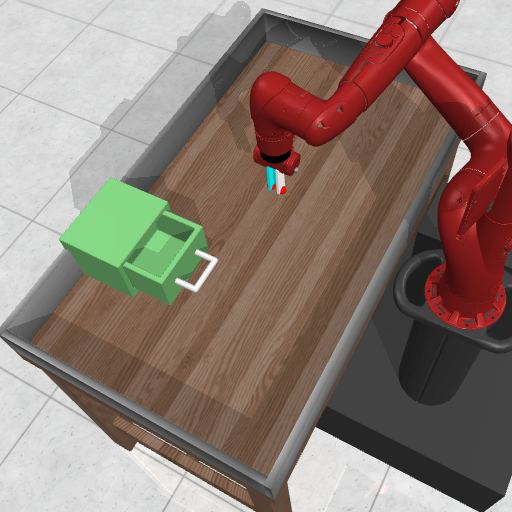}
    \caption{\footnotesize \textbf{Multitask Drawer Environment.}: We apply cross entropy method (CEM) optimization on the LAEO dynamics model trained only on the data from the drawer close task to solve six different tasks: close, half-closed, open, reach (near), reach (medium), and reach (far). .
    }
    \label{fig:multitask_env}
    \vspace{-0.5em}
\end{figure}

The \texttt{Half-closed} task requires the agent to push the drawer from an open position into a halfway closed position. The \texttt{Open} task requires the agent to pull the drawer from a closed position into an open position. The \texttt{Close} task is the same as in the original \texttt{SawyerDrawerClose} environment, and requires the agent to push the drawer from an opened position into a closed position. 
The three reaching tasks, \texttt{Reach-near}, \texttt{Reach-medium}, \texttt{Reach-far}, require the agent to reach the end-effector to a three different target positions. The tasks are visualized in Figure \ref{fig:multitask_env}.

For these experiments, we load the final checkpoint of a critic network from the previous set of experiments (Comparison to Goal-Conditioned RL), and select actions by using cross entropy method (CEM) optimization on the critic network. By using CEM optimization, we do not need to train a separate policy network for each of the tasks.
Since the LAEO dynamics model is a multistep dynamics model (meaning that the model predicts whether a goal state will be reached sometime in the future and not just at the subsequent timestep) we are able to use CEM directly with the dynamics model. Specifically, for each task, we collect a success example using a scripted policy from~\citet{yu2020metaworld}. Then, at each environment timestep $t$, we condition the LAEO dynamics model on the success example, and then run CEM to choose an action that maximizes the output of the dynamics network. 
At each timestep, we perform 10 iterations of CEM, using a population size of $10,000$ and an elite population size of $2,000$.
Results are averaged across eight random seeds.

\end{document}